# Hinge-loss Markov Random Fields: Convex Inference for Structured Prediction

**Stephen H. Bach**   **Bert Huang**   **Ben London**   **Lise Getoor**
Computer Science Dept.
University of Maryland
College Park, MD, 20742, USA

## Abstract

Graphical models for structured domains are powerful tools, but the computational complexities of combinatorial prediction spaces can force restrictions on models, or require approximate inference in order to be tractable. Instead of working in a combinatorial space, we use hinge-loss Markov random fields (HL-MRFs), an expressive class of graphical models with log-concave density functions over continuous variables, which can represent confidences in discrete predictions. This paper demonstrates that HL-MRFs are general tools for fast and accurate structured prediction. We introduce the first inference algorithm that is both scalable and applicable to the full class of HL-MRFs, and show how to train HL-MRFs with several learning algorithms. Our experiments show that HL-MRFs match or surpass the predictive performance of state-of-the-art methods, including discrete models, in four application domains.

## 1 INTRODUCTION

The study of probabilistic modeling in structured and relational domains primarily focuses on predicting discrete variables [12, 19, 24]. However, except for some isolated cases, the combinatorial nature of discrete, structured prediction spaces requires concessions: most notably, for inference algorithms to be tractable, they must be relaxed or approximate (e.g., [19, 22, 25]), or the model's structure must be restricted (e.g., [2, 8]). Broecheler et al. [6] introduced a class of models for continuous variables that has the potential to combine fast and exact inference with the expressivity of discrete graphical models, but this potential has not been well explored. Now called *hinge-loss Markov random fields* (HL-MRFs) [3], these models are analogous to discrete Markov random fields, except that random variables are continuous valued in the unit interval [0,1], and potentials are linear or squared hinge-loss functions.

In this work, we demonstrate that HL-MRFs are powerful tools for structured prediction by producing state-of-the-art performance in a number of domains. We are the first to leverage some of the most powerful features of HL-MRFs, such as squared potentials, which we show produce better results on multiple tasks. We show that HL-MRFs are well-suited to structured prediction for the following reasons. They are expressive, interpretable, and easily defined using the modeling language *probabilistic soft logic* (PSL) [6, 13]. Further, continuous variables are useful both for modeling continuous data as well as for expressing confidences in discrete predictions. Confidences are desirable for the same reason that practitioners often prefer marginal probabilities to the single most probable discrete prediction. Finally, HL-MRFs have log-concave density functions, so finding an exact *most probable explanation* (MPE) for a given input is a convex optimization, and therefore exactly solvable in polynomial time.

Our specific contributions include the following. First, we introduce a new, fast algorithm for MPE inference in HL-MRFs, which is the first to be both scalable and applicable to the full class of HL-MRFs. Second, we show how to train HL-MRFs with several learning algorithms. Third, we show empirically that these advances enable HL-MRFs to tackle a diverse set of relational and structured prediction tasks, providing state-of-the-art performance on collective classification, social-trust prediction, collaborative filtering, and image reconstruction. In particular, we show that HL-MRFs can outperform their discrete counterparts, as well as other leading methods.

## 1.1 RELATED WORK

*Probabilistic soft logic* (PSL) [6, 13] is a declarative language for defining templated HL-MRFs. Its development was partially motivated by the need for rich models of continuous similarity values, for use in tasks such as entity resolution, collective classification, and ontology alignment. PSL is in a family of systems for defining templated, relational probabilistic models that includes, for instance, *Markov logic networks* [19], *relational dependency networks* [17], and *relational Markov networks* [23]. In our experiments, we compare against Markov logic networks, which use a first-order syntax similar to PSL's to build discrete probabilistic models.

MPE inference algorithms for HL-MRFs solve a constrained, convex optimization. A standard approach for general, constrained, convex optimization is to use an interior-point method, which Broecheler et al. [6] use. While theoretically efficient, the practical running time of interior-point optimizers quickly becomes cumbersome for large problems. For discrete graphical models, recent advances use *consensus optimization* to obtain fast, approximate MPE inference algorithms [5, 15, 16]. Bach et al. [3] recently developed an analogous algorithm for exact MPE inference in HL-MRFs that produced a significant improvement in running time over interior-point methods, though it was limited to pairwise potentials and constraints, and cost considerably more computation to optimize over squared potentials.

The learning methods we adapt for HL-MRFs are standard approaches for learning parameters of probabilistic models. In particular, our adaptations are analogous to previous learning algorithms for relational and structured models using approximate maximum-likelihood or maximum-pseudolikelihood estimation [6, 14, 17, 19] and large-margin estimation [11, 12, 24].

## 2 HINGE-LOSS MARKOV RANDOM FIELDS

Hinge-loss Markov random fields (HL-MRFs) are a general class of conditional, continuous probabilistic models [3]. HL-MRFs are log-linear models whose features are hinge-loss functions of the variable states. Through constructions based on *soft logic*, hinge-loss potentials can be used to model generalizations of logical conjunction and implication. HL-MRFs can be defined using the modeling language *probabilistic soft logic* (PSL) [6, 13], making these powerful models interpretable, flexible, and expressive. In this section, we formally present constrained hinge-loss energy functions, HL-MRFs, and briefly review PSL.

**Definition 1.** *Let* $\mathbf{Y} = (Y_1, \ldots, Y_n)$ *be a vector of* $n$ *variables and* $\mathbf{X} = (X_1, \ldots, X_{n'})$ *a vector of* $n'$ *variables with joint domain* $\mathbf{D} = [0,1]^{n+n'}$. *Let* $\phi = (\phi_1, \ldots, \phi_m)$ *be* $m$ *continuous potentials of the form*

$$\phi_j(\mathbf{Y}, \mathbf{X}) = [\max\{\ell_j(\mathbf{Y}, \mathbf{X}), 0\}]^{p_j}$$

*where* $\ell_j$ *is a linear function of* $\mathbf{Y}$ *and* $\mathbf{X}$ *and* $p_j \in \{1,2\}$. *Let* $C = (C_1, \ldots, C_r)$ *be linear constraint functions associated with index sets denoting equality constraints* $\mathcal{E}$ *and inequality constraints* $\mathcal{I}$, *which define the feasible set*

$$\tilde{\mathbf{D}} = \left\{ \mathbf{Y}, \mathbf{X} \in \mathbf{D} \;\middle|\; \begin{array}{l} C_k(\mathbf{Y}, \mathbf{X}) = 0, \forall k \in \mathcal{E} \\ C_k(\mathbf{Y}, \mathbf{X}) \geq 0, \forall k \in \mathcal{I} \end{array} \right\}.$$

*For* $\mathbf{Y}, \mathbf{X} \in \tilde{\mathbf{D}}$, *given a vector of nonnegative free parameters, i.e., weights,* $\lambda = (\lambda_1, \ldots, \lambda_m)$, *a constrained hinge-loss energy function* $f_\lambda$ *is defined as*

$$f_\lambda(\mathbf{Y}, \mathbf{X}) = \sum_{j=1}^{m} \lambda_j \phi_j(\mathbf{Y}, \mathbf{X}).$$

**Definition 2.** *A hinge-loss Markov random field* $P$ *over random variables* $\mathbf{Y}$ *and conditioned on random variables* $\mathbf{X}$ *is a probability density defined as follows: if* $\mathbf{Y}, \mathbf{X} \notin \tilde{\mathbf{D}}$, *then* $P(\mathbf{Y}|\mathbf{X}) = 0$; *if* $\mathbf{Y}, \mathbf{X} \in \tilde{\mathbf{D}}$, *then*

$$P(\mathbf{Y}|\mathbf{X}) = \frac{1}{Z(\lambda)} \exp\left[-f_\lambda(\mathbf{Y}, \mathbf{X})\right],$$

*where* $Z(\lambda) = \int_\mathbf{Y} \exp\left[-f_\lambda(\mathbf{Y}, \mathbf{X})\right]$.

Thus, MPE inference is equivalent to finding the minimizer of the convex energy $f_\lambda$.

The potentials and weights can be grouped together into *templates*, which can be used to define general classes of HL-MRFs that are parameterized by the input data. Let $\mathcal{T} = (t_1, \ldots, t_s)$ denote a vector of templates with associated weights $\Lambda = (\Lambda_1, \ldots, \Lambda_s)$. We partition the potentials by their associated templates and let $\Phi_q(\mathbf{Y}, \mathbf{X}) = \sum_{j \in t_q} \phi_j(\mathbf{Y}, \mathbf{X})$ for all $t_q \in \mathcal{T}$. In the HL-MRF, the weight of the $j$'th hinge-loss potential is set to the weight of the template from which it was derived, i.e., $\lambda_j = \Lambda_q$, for each $j \in t_q$.

Probabilistic soft logic [6, 13] provides a natural interface to represent hinge-loss potential templates using logical rules. In particular, a logical conjunction of Boolean variables $X \wedge Y$ can be generalized to continuous variables using the hinge function $\max\{X + Y - 1, 0\}$, which is known as the *Lukasiewicz t-norm*. Disjunction $X \vee Y$ is relaxed to $\min\{X + Y, 1\}$, and negation $\neg X$ to $1 - X$. PSL allows modelers to design rules that, given data, ground out substitutions for logical terms. The groundings of a template define hinge-loss potentials that share the same weight and have the form one minus the truth value of the ground rule. We defer to Broecheler et al. [6] and Kimmig et al. [13] for details on PSL.

To further demonstrate this templating, consider the task of predicting who trusts whom in a social network. Let the network contain three individuals: $A$, $B$, and $C$. We can design an HL-MRF to include potentials that encode the belief that trust is transitive (which is a rule we use in our experiments). Let the variable $Y_{A,B}$ represent how much $A$ trusts $B$, and similarly so for $Y_{B,C}$ and $Y_{A,C}$. Then the potential

$$\phi(\mathbf{Y}, \mathbf{X}) = [\max\{Y_{A,B} + Y_{B,C} - Y_{A,C} - 1, 0\}]^p$$

is equivalent to one minus the truth value of the Boolean formula $Y_{A,B} \wedge Y_{B,C} \rightarrow Y_{A,C}$ when $Y_{A,B}, Y_{B,C}, Y_{A,C} \in \{0, 1\}$. When they are allowed to take on their full range $[0, 1]$, the potential is a convex relaxation of the implication. An HL-MRF with this potential function assigns higher probability to variable states that satisfy the logical implication above, which can occur to varying degrees in the continuous domain. Given a social network with more than these three individuals, PSL can ground out possible substitutions for the roles of $A$, $B$, and $C$ to generate potential functions for each substitution, thus defining the full, ground HL-MRF.

HL-MRFs support a few additional components useful for modeling. The constraints in Definition 1 allow the encoding of functional modeling requirements, which is useful, e.g., when variables correspond to mutually exclusive labels, and thus should sum to one. The exponent parameter $p_j$ allows flexibility in the shape of the hinge, affecting the sharpness of the penalty for violating the logical implication. Setting $p_j$ to 1 penalizes violation linearly with the amount the implication is unsatisfied, while setting $p_j$ to 2 penalizes small violations much less. In effect, some linear potentials overrule others, while the influences of squared potentials are averaged together.

## 3 MPE INFERENCE

MPE inference for HL-MRFs requires finding a feasible assignment that minimizes $f_\lambda$. Performing MPE inference quickly is crucial, especially because weight learning often requires performing inference many times with different weights (as we discuss in Section 4). Here, HL-MRFs have a distinct advantage over general discrete models, since minimizing $f_\lambda$ is a convex optimization rather than a combinatorial one. In this section, we detail a new, faster MPE inference algorithm for HL-MRFs.

Bach et al. [3] showed how to minimize $f_\lambda$ with a consensus-optimization algorithm, based on the *alternating-direction method of multipliers* (ADMM) [5]. The algorithm works by creating local copies of the variables in each potential and constraint, constraining them to be equal to the original variables, and relaxing those equality constraints to make independent subproblems. By solving the subproblems repeatedly and averaging the results, the algorithm reaches a consensus on the best values of the original variables, also called the *consensus variables*. This procedure is guaranteed to converge to the global minimizer of $f_\lambda$. See [3] and [5] for more details on consensus optimization and ADMM.

This previous consensus-optimization approach to MPE inference works well for linear potentials with at most two unobserved variables, and empirical evidence suggests it scales linearly with the size of the problem [3]. However, it is restricted to pairwise potentials and constraints, and requires an interior-point method as a subroutine to solve subproblems induced by squared potentials. Because of the embedded interior-point method, its running time can increase roughly 100-fold with squared potentials [3].

We improve the algorithm of Bach et al. [3] by reformulating the optimization to enforce $\mathbf{Y} \in [0, 1]^n$ only on the consensus variables, not the local copies. This form of consensus optimization is described in greater detail by Boyd et al. [5]. The result is that, in our algorithm, the potentials and constraints are not restricted to a certain number of unknowns, and the subproblems can all be solved quickly using simple linear algebra.

Algorithm 1 gives pseudocode for our new algorithm. It starts by initializing local copies of the variables that appear in each potential and constraint, along with a corresponding Lagrange multiplier for each copy. Then, until convergence, it iteratively updates Lagrange multipliers and solves suproblems induced by the HL-MRF's potentials and constraints. If the subproblem is induced by a potential, it sets the local variable copies to a balance between the minimizer of the potential and the emerging consensus. Eventually the Lagrange multipliers will enforce agreement between the local copies and the consensus. If instead the subproblem is induced by a constraint in the HL-MRF, the algorithm projects the consensus variables' current values to that constraint's feasible region. The consensus variables are updated at each iteration based on the values of their local copies and corresponding Lagrange multipliers, and clipped to [0,1].

We take the same basic approach as Bach et al. [3] to solve the subproblems. We first try to find a minimizer on either side of the hinge (where $\ell_j(\mathbf{Y}, \mathbf{X})$ is either positive or negative) before projecting onto the plane defined by $\ell_j(\mathbf{Y}, \mathbf{X}) = 0$. Without the constraints on

**Algorithm 1** MPE Inference for HL-MRFs

**Input:** HL-MRF($\mathbf{Y}, \mathbf{X}, \phi, \lambda, C, \mathcal{E}, \mathcal{I}$), $\rho > 0$

Initialize $\mathbf{y}_j$ as copies of the variables $\mathbf{Y}_j$ that appear in $\phi_j$, $j = 1, \ldots, m$

Initialize $\mathbf{y}_{k+m}$ as copies of the variables $\mathbf{Y}_{k+m}$ that appear in $C_k$, $k = 1, \ldots, r$

Initialize Lagrange multipliers $\boldsymbol{\alpha}_i$ corresponding to variable copies $\mathbf{y}_i$, $i = 1, \ldots, m+r$

**while** not converged **do**
  **for** $j = 1, \ldots, m$ **do**
    $\boldsymbol{\alpha}_j \leftarrow \boldsymbol{\alpha}_j + \rho(\mathbf{y}_j - \mathbf{Y}_j)$
    $\mathbf{y}_j \leftarrow \mathbf{Y}_j - \boldsymbol{\alpha}_j/\rho$
    **if** $\ell_j(\mathbf{y}_j, \mathbf{X}) > 0$ **then**
      $\mathbf{y}_j \leftarrow \arg\min_{\mathbf{y}_j} \left[ \begin{array}{l} \lambda_j[\ell_j(\mathbf{y}_j, \mathbf{X})]^{p_j} \\ +\frac{\rho}{2}\|\mathbf{y}_j - \mathbf{Y}_j + \frac{1}{\rho}\boldsymbol{\alpha}_j\|_2^2 \end{array} \right]$
      **if** $\ell_j(\mathbf{y}_j, \mathbf{X}) < 0$ **then**
        $\mathbf{y}_j \leftarrow \text{Proj}_{\ell_j=0}(\mathbf{Y}_j)$
      **end if**
    **end if**
  **end for**

  **for** $k = 1, \ldots, r$ **do**
    $\boldsymbol{\alpha}_{k+m} \leftarrow \boldsymbol{\alpha}_{k+m} + \rho(\mathbf{y}_{k+m} - \mathbf{Y}_{k+m})$
    $\mathbf{y}_{k+m} \leftarrow \text{Proj}_{C_k}(\mathbf{Y}_{k+m})$
  **end for**

  **for** $i = 1, \ldots, n$ **do**
    $Y_i \leftarrow \frac{1}{|\texttt{copies}(Y_i)|} \sum_{y_c \in \texttt{copies}(Y_i)} \left( y_c + \frac{\alpha_c}{\rho} \right)$
    Clip $Y_i$ to [0,1]
  **end for**
**end while**

---

the local variable copies, finding these minimizers and projections is much simpler. Solving for the constraint subproblems is now also simpler, requiring just a projection onto a hyperplane or a halfspace.

Our algorithm retains all of the benefits of the original MPE inference algorithm while removing all restrictions on the numbers of unknowns in the potentials and constraints, making MPE inference fast with both linear and squared potentials. Another advantage of consensus optimization for MPE inference is that it is easy to warm start. Warm starting provides significant efficiency gains when inference is repeated on the same HL-MRF with small changes in the weights, as often occurs during weight learning.

## 4 WEIGHT LEARNING

In this section, we present three weight learning methods for HL-MRFs, each with a different objective function, two of which are new for learning HL-MRFs. The first method, introduced by Broecheler et al. [6], performs approximate maximum-likelihood estimation using MPE inference to approximate the gradient of the log-likelihood. The second method maximizes the pseudolikelihood. The third method finds a large-margin solution, preferring weights that discriminate the ground truth from other states. We describe below how to apply these learning strategies to HL-MRFs.

### 4.1 MAXIMUM-LIKELIHOOD ESTIMATION

The canonical approach for learning parameters $\Lambda$ is to maximize the log-likelihood of training data. The partial derivative of the log-likelihood with respect to a parameter $\Lambda_q$ is

$$\frac{\partial \log p(\mathbf{Y}|\mathbf{X})}{\partial \Lambda_q} = \mathbb{E}_\Lambda \left[ \Phi_q(\mathbf{Y}, \mathbf{X}) \right] - \Phi_q(\mathbf{Y}, \mathbf{X}) ,$$

where $\mathbb{E}_\Lambda$ is the expectation under the distribution defined by $\Lambda$. The voted perceptron algorithm [7] optimizes $\Lambda$ by taking steps of fixed length in the direction of the gradient, then averaging the points after all steps. Any step that is outside the feasible region is projected back before continuing. For a smoother ascent, it is often helpful to divide the $q$-th component of the gradient by the number of groundings $|t_q|$ of the $q$'th template [14], which we do in our experiments. Computing the expectation is intractable, so we use a common approximation: the values of the potential functions at the most probable setting of $\mathbf{Y}$ with the current parameters [6].

### 4.2 MAXIMUM-PSEUDOLIKELIHOOD ESTIMATION

Since exact maximum likelihood estimation is intractable in general, we can instead perform *maximum-pseudolikelihood estimation* (MPLE) [4], which maximizes the likelihood of each variable conditioned on all other variables, i.e.,

$$P^*(\mathbf{Y}|\mathbf{X}) = \prod_{i=1}^n P^*(Y_i|\text{MB}(Y_i))$$
$$= \prod_{i=1}^n \frac{1}{Z(\lambda, Y_i)} \exp\left[-f_\lambda^i(Y_i, \mathbf{Y}, \mathbf{X})\right];$$
$$Z(\lambda, Y_i) = \int_{Y_i} \exp\left[-f_\lambda^i(Y_i, \mathbf{Y}, \mathbf{X})\right];$$
$$f_\lambda^i(Y_i, \mathbf{Y}, \mathbf{X}) = \sum_{j:i \in \phi_j} \lambda_j \phi_j\left(\{Y_i \cup \mathbf{Y}_{\setminus i}\}, \mathbf{X}\right).$$

Here, $i \in \phi_j$ means that $Y_i$ is involved in $\phi_j$, and $\text{MB}(Y_i)$ denotes the *Markov blanket* of $Y_i$—that is, the set of variables that co-occur with $Y_i$ in any po-

tential function. The partial derivative of the log-pseudolikelihood with respect to $\Lambda_q$ is

$$\frac{\partial \log P^*(Y|X)}{\partial \Lambda_q} = \sum_{i=1}^{n} \mathbb{E}_{Y_i|MB} \left[ \sum_{j \in t_q : i \in \phi_j} \phi_j(\mathbf{Y}, \mathbf{X}) \right] - \Phi_j(\mathbf{Y}, \mathbf{X}).$$

Computing the pseudolikelihood gradient does not require inference and takes time linear in the size of $\mathbf{Y}$. However, the integral in the above expectation does not readily admit a closed-form antiderivative, so we approximate the expectation. When a variable in unconstrained, the domain of integration is a one-dimensional interval on the real number line, so Monte Carlo integration quickly converges to an accurate estimate of the expectation.

We can also apply MPLE when the constraints are not too interdependent. For example, for linear equality constraints over disjoint groups of variables (e.g., variable sets that must sum to 1.0), we can block-sample the constrained variables by sampling uniformly from a simplex. These types of constraints are often used to represent mutual exclusivity of classification labels. We can compute accurate estimates quickly because these blocks are typically low-dimensional.

### 4.3 LARGE-MARGIN ESTIMATION

A different approach to learning drops the probabilistic interpretation of the model and views HL-MRF inference as a prediction function. Large-margin estimation (LME) shifts the goal of learning from producing accurate probabilistic models to instead producing accurate MPE predictions. The learning task is then to find the weights $\Lambda$ that provide high-accuracy structured predictions. We describe in this section a large-margin method based on the cutting-plane approach for *structural support vector machines* (SVMs) [12].

The intuition behind large-margin structured prediction is that the ground-truth state should have energy lower than any alternate state by a large margin. In our setting, the output space is continuous, so we parameterize this margin criterion with a continuous loss function. For any valid output state $\tilde{\mathbf{Y}}$, a large-margin solution should satisfy:

$$f_\lambda(\mathbf{Y}, \mathbf{X}) \leq f_\lambda(\tilde{\mathbf{Y}}, \mathbf{X}) - L(\mathbf{Y}, \tilde{\mathbf{Y}}), \forall \tilde{\mathbf{Y}},$$

where the decomposable loss function $L(\mathbf{Y}, \tilde{\mathbf{Y}}) = \sum_i L(Y_i, \tilde{Y}_i)$ measures the disagreement between a state $\tilde{\mathbf{Y}}$ and the training label state $\mathbf{Y}$. We define $L$ as the $\ell_1$ distance. Since we do not expect all problems to be perfectly separable, we relax this constraint with a penalized slack $\xi$. We obtain a convex learning objective for a large-margin solution

$$\min_{\Lambda \geq 0} \quad \frac{1}{2}||\Lambda||^2 + C\xi$$
$$\text{s.t.} \quad \Lambda^\top (\Phi(\mathbf{Y}, \mathbf{X}) - \Phi(\tilde{\mathbf{Y}}, \mathbf{X})) \leq -L(\mathbf{Y}, \tilde{\mathbf{Y}}) + \xi, \forall \mathbf{Y},$$

where $\Phi(\mathbf{Y}, \mathbf{X}) = (\Phi_1(\mathbf{Y}, \mathbf{X}), \ldots, \Phi_s(\mathbf{Y}, \mathbf{X}))$. This formulation is analogous to the margin-rescaling approach by Joachims et al. [12]. Though such a structured objective is natural and intuitive, its number of constraints is the cardinality of the output space, which here is infinite. Following their approach, we optimize subject to the infinite constraint set using a *cutting-plane algorithm*: we greedily grow a set $K$ of constraints by iteratively adding the worst-violated constrain given by a *separation oracle*, then updating $\Lambda$ subject to the current constraints. The goal of the cutting-plane approach is to efficiently find the set of active constraints at the solution for the full objective, without having to enumerate the infinite inactive constraints. The worst-violated constraint is

$$\arg\min_{\tilde{\mathbf{Y}}} \Lambda^\top \Phi(\tilde{\mathbf{Y}}, \mathbf{X}) - L(\mathbf{Y}, \tilde{\mathbf{Y}}).$$

The separation oracle performs loss-augmented inference by adding additional loss-augmenting potentials to the HL-MRF. For ground truth in $\{0, 1\}$, these loss-augmenting potentials are also examples of hinge-losses, and thus adding them simply creates an augmented HL-MRF. The worst-violated constraint is then computed as standard inference on the loss-augmented HL-MRF. However, ground-truth variables in the interior $(0, 1)$ cause any distance-based loss to be concave, which require the separation oracle to solve a non-convex objective. For interior ground truth values, we use the *difference of convex functions algorithm* [1] to find a local optimum. Since the concave portion of the loss-augmented inference objective pivots around the ground truth value, the subgradients are 1 or $-1$, depending on whether the current value is greater than the ground truth. We simply choose an initial direction for interior labels by rounding, and flip the direction of the subgradients for variables whose solution states are not in the interval corresponding to the subgradient direction until convergence.

Given a set $K$ of constraints, we solve the SVM objective as in the primal form $\min_{\Lambda \geq 0} \frac{1}{2}||\Lambda||^2 + C\xi$ s.t. $K$. We then iteratively invoke the separation oracle to find the worst-violated constraint. If this new constraint is not violated, or its violation is within numerical tolerance, we have found the max-margin solution. Otherwise, we add the new constraint to $K$, and repeat.

One fact of note is that the large-margin criterion always requires a little slack for squared HL-MRFs. Since the squared hinge potential is quadratic and the loss is linear, there always exists a small enough distance from the ground truth such that an absolute (i.e., linear) distance is greater than the squared distance. In these cases, the slack parameter trades off between the peakedness of the learned quadratic energy function and the margin criterion.

## 5 EXPERIMENTS

To demonstrate the flexibility and effectiveness of HL-MRFs, we test them on four diverse learning tasks: collective classification, social-trust prediction, preference prediction, and image reconstruction. [1] Each of these experiments represents a problem domain that is best solved with relational learning approaches because structure is a critical component of their problems. The experiments show that HL-MRFs perform as well as or better than state-of-the-art approaches.

For these diverse tasks, we compare against a number of competing methods. For collective classification and social-trust prediction, we compare HL-MRFs to discrete Markov random fields (MRFs). We construct them with Markov logic networks (MLNs) [19], which template discrete MRFs using logical rules similarly to PSL for HL-MRFs. We perform inference in discrete MRFs using 2500 rounds of the sampling algorithm *MC-Sat* (500 of which are burn in), and we find approximate MPE states during MLE learning using the search algorithm *MaxWalkSat* [19]. For collaborative filtering, a task that is inherently continuous and nontrivial to encode in discrete logic, we compare against *Bayesian probabilistic matrix factorization* [20]. Finally, for image reconstruction, we run the same experimental setup as Poon and Domingos [18] and compare against the results they report, which include tests using *sum product networks*, *deep belief networks*, and *deep Boltzmann machines*.

When appropriate, we evaluate statistical significance using a paired t-test with rejection threshold 0.01. We omit variance statistics to save space and only report the average and statistical significance. We describe the HL-MRFs used for our experiments using the PSL rules that define them. To investigate the differences between linear and squared potentials we use both in our experiments. HL-MRF-L refers to a model with all linear potentials and HL-MRF-Q to one with all squared potentials. When training with MLE and MPLE, we use 100 steps of voted perceptron and a step size of 1.0 (unless otherwise noted), and for LME we set $C = 0.1$. We experimented with various settings, but the scores of HL-MRFs and discrete MRFs were not sensitive to changes.

---

[1] All code is available at http://psl.umiacs.umd.edu.

Table 1: Average accuracy of classification by HL-MRFs and discrete MRFs. Scores statistically equivalent to the best scoring method are typed in bold.

|  | Citeseer | Cora |
|---|---|---|
| HL-MRF-Q (MLE) | **0.729** | **0.816** |
| HL-MRF-Q (MPLE) | **0.729** | **0.818** |
| HL-MRF-Q (LME) | 0.683 | 0.789 |
| HL-MRF-L (MLE) | **0.724** | 0.802 |
| HL-MRF-L (MPLE) | **0.729** | **0.808** |
| HL-MRF-L (LME) | 0.695 | 0.789 |
| MRF (MLE) | 0.686 | 0.756 |
| MRF (MPLE) | 0.715 | 0.797 |
| MRF (LME) | 0.687 | 0.783 |

### 5.1 COLLECTIVE CLASSIFICATION

When classifying documents, links between those documents—such as hyperlinks, citations, or co-authorship—provide extra signal beyond the local features of individual documents. Collectively predicting document classes with these links tends to improve accuracy [21]. We classify documents in citation networks using data from the Cora and Citeseer scientific paper repositories. The Cora data set contains 2,708 papers in seven categories, and 5,429 directed citation links. The Citeseer data set contains 3,312 papers in six categories, and 4,591 directed citation links.

The prediction task is, given a set of seed documents whose labels are observed, to infer the remaining document classes by propagating the seed information through the network. For each of 20 runs, we split the data sets 50/50 into training and testing partitions, and seed half of each set. To predict discrete categories with HL-MRFs we predict the category with the highest predicted value.

We compare HL-MRFs to discrete MRFs on this task. We construct both using the same logical rules, which simply encode the tendency for a class to propagate across citations. For each category $C_i$, we have two separate rules for each direction of citation:

$$\text{LABEL}(A, C_i) \wedge \text{CITES}(A, B) \Rightarrow \text{LABEL}(B, C_i),$$
$$\text{LABEL}(A, C_i) \wedge \text{CITES}(B, A) \Rightarrow \text{LABEL}(B, C_i).$$

Table 1 lists the results of this experiment. HL-MRFs are the most accurate predictors on both data sets. We also note that both variants of HL-MRFs are much faster than discrete MRFs. See Table 3 for average inference times on five folds.

Table 2: Average area under ROC and precision-recall curves of social-trust prediction by HL-MRFs and discrete MRFs. Scores statistically equivalent to the best scoring method by metric are typed in bold.

|                  | ROC   | P-R (+) | P-R (−) |
|------------------|-------|---------|---------|
| HL-MRF-Q (MLE)   | **0.822** | **0.978** | 0.452 |
| HL-MRF-Q (MPLE)  | **0.832** | **0.979** | 0.482 |
| HL-MRF-Q (LME)   | **0.814** | **0.976** | 0.462 |
| HL-MRF-L (MLE)   | 0.765 | 0.965 | 0.357 |
| HL-MRF-L (MPLE)  | 0.757 | 0.963 | 0.333 |
| HL-MRF-L (LME)   | 0.783 | 0.967 | **0.453** |
| MRF (MLE)        | 0.655 | 0.942 | 0.270 |
| MRF (MPLE)       | 0.725 | 0.963 | 0.298 |
| MRF (LME)        | 0.795 | **0.973** | **0.441** |

## 5.2 SOCIAL-TRUST PREDICTION

An emerging problem in the analysis of online social networks is the task of inferring the level of trust between individuals. Predicting the strength of trust relationships can provide useful information for viral marketing, recommendation engines, and internet security. HL-MRFs with linear potentials have recently been applied by Huang et al. [10] to this task, showing superior results with models based on sociological theory. We reproduce their experimental setup using their sample of the signed Epinions trust network, in which users indicate whether they trust or distrust other users. We perform eight-fold cross-validation. In each fold, the prediction algorithm observes the entire unsigned social network and all but 1/8 of the trust ratings. We measure prediction accuracy on the held-out 1/8. The sampled network contains 2,000 users, with 8,675 signed links. Of these links, 7,974 are positive and only 701 are negative.

We use a model based on the social theory of *structural balance*, which suggests that social structures are governed by a system that prefers triangles that are considered balanced. Balanced triangles have an odd number of positive trust relationships; thus, considering all possible directions of links that form a triad of users, there are sixteen logical implications of the form

$$\text{Trusts}(A, B) \land \text{Trusts}(B, C) \Rightarrow \text{Trusts}(A, C).$$

Huang et al. [10] list all sixteen of these rules, a reciprocity rule, and a prior in their *Balance-Recip* model, which we omit to save space.

Since we expect some of these structural implications to be more or less accurate, learning weights for these rules provides better models. Again, we use these rules to define HL-MRFs and discrete MRFs, and we train them using various learning algorithms. We compute three metrics: the area under the receiver operating characteristic (ROC) curve, and the areas under the precision-recall curves for positive trust and negative trust. On all three metrics, HL-MRFs with squared potentials score significantly higher. The differences among the learning methods for squared HL-MRFs are insignificant, but the differences among the models is statistically significant for the ROC metric. For area under the precision-recall curve for positive trust, discrete MRFs trained with LME are statistically tied with the best score, and both HL-MRF-L and discrete MRFs trained with LME are statistically tied with the best area under the precision-recall curve for negative trust. The results are listed in Table 2.

Table 3: Average inference times (reported in seconds) of single-threaded HL-MRFs and discrete MRFs.

|           | Citeseer | Cora   | Epinions |
|-----------|----------|--------|----------|
| HL-MRF-Q  | 0.42     | 0.70   | 0.32     |
| HL-MRF-L  | 0.46     | 0.50   | 0.28     |
| MRF       | 110.96   | 184.32 | 212.36   |

Though the random fold splits are not the same, using the same experimental setup, Huang et al. [10] also scored the precision-recall area for negative trust of standard trust prediction algorithms EigenTrust and TidalTrust, which scored 0.131 and 0.130, respectively. The logical models based on structural balance that we run here are significantly more accurate, and HL-MRFs more than discrete MRFs.

Table 3 lists average inference times on five folds of three prediction tasks: Cora, Citeseer, and Epinions. We implemented each method in Java. Both HL-MRF-Q and HL-MRF-L are much faster than discrete MRFs. This illustrates an important difference between performing structured prediction via convex inference versus sampling in a discrete prediction space: using our MPE inference algorithm is much faster.

## 5.3 PREFERENCE PREDICTION

Preference prediction is the task of inferring user attitudes (often quantified by ratings) toward a set of items. This problem is naturally structured, since a user's preferences are often interdependent, as are an item's ratings. *Collaborative filtering* is the task of predicting unknown ratings using only a subset of observed ratings. Methods for this task range from simple nearest-neighbor classifiers to complex latent factor models. To illustrate the versatility of HL-MRFs, we design a simple, interpretable collaborative filtering model for predicting humor preferences. We test this model on the Jester dataset, a repository of ratings from 24,983 users on a set of 100 jokes [9]. Each joke is rated on a scale of $[-10, +10]$, which we normalize to $[0, 1]$. We sample a random 2,000 users from the set of those who rated all 100 jokes, which we then split into 1,000 train and 1,000 test users. From each train and test matrix, we sample a random 50% to use as the observed features $\mathbf{X}$; the remaining ratings are treated as the variables $\mathbf{Y}$.

Table 4: Normalized mean squared/absolute errors (NMSE/NMAE) for preference prediction using the Jester dataset. The lowest errors are typed in bold.

|                  | NMSE   | NMAE   |
|------------------|--------|--------|
| HL-MRF-Q (MLE)   | 0.0554 | 0.1974 |
| HL-MRF-Q (MPLE)  | 0.0549 | 0.1953 |
| HL-MRF-Q (LME)   | 0.0738 | 0.2297 |
| HL-MRF-L (MLE)   | 0.0578 | 0.2021 |
| HL-MRF-L (MPLE)  | 0.0535 | 0.1885 |
| HL-MRF-L (LME)   | 0.0544 | **0.1875** |
| BPMF             | **0.0501** | **0.1832** |

Our HL-MRF model uses an item-item similarity rule:

$$\text{SimRate}(J_1, J_2) \wedge \text{Likes}(U, J_1) \Rightarrow \text{Likes}(U, J_2)$$

where $J_1, J_2$ are jokes and $U$ is a user; the predicate LIKES indicates the degree of preference (i.e., rating value); and SIMRATE measures the mean-adjusted cosine similarity between the observed ratings of two jokes. We also include rules to enforce that $\text{Likes}(U, J)$ concentrates around the observed average rating of user $U$ and item $J$, and the global average.

We compare our HL-MRF model to a current state-of-the-art latent factors model, *Bayesian probabilistic matrix factorization* (BPMF) [20]. BPMF is a fully Bayesian treatment and, as such, is considered "parameter-free"; the only parameter that must be specified is the rank of the decomposition. For our experiments, we use Xiong et al.'s code [2010]. Since BPMF does not train a model, we allow BPMF to use all of the training matrix during the prediction phase.

Table 4 lists the normalized mean squared error (NMSE) and normalized mean absolute error (NMAE), averaged over 10 random splits. Though BPMF produces the best scores, the improvement over HL-MRF-L (LME) is not significant in NMAE.

## 5.4 IMAGE RECONSTRUCTION

Digital image reconstruction requires models that understand how pixels relate to each other, such that when some pixels are unobserved, the model can infer their values from parts of the image that are observed. We construct pixel-grid HL-MRFs for image reconstruction. We test these models using the experimental setup of Poon and Domingos [18]: we reconstruct images from the Olivetti face data set and the Caltech101 face category. The Olivetti data set contains 400 images, 64 pixels wide and tall, and the Caltech101 face category contains 435 examples of faces, which we crop to the center 64 by 64 patch, as was done by Poon and Domingos [18]. Following their experimental setup, we hold out the last fifty images and predict either the left half of the image or the bottom half.

The HL-MRFs in this experiment are much more complex than the ones in our other experiments because we allow each pixel to have its own weight for the following rules, which encode agreement or disagreement between neighboring pixels:

$$\text{Bright}(P_{ij}, I) \wedge \text{North}(P_{ij}, Q) \Rightarrow \text{Bright}(Q, I),$$
$$\text{Bright}(P_{ij}, I) \wedge \text{North}(P_{ij}, Q) \Rightarrow \neg\text{Bright}(Q, I),$$
$$\neg\text{Bright}(P_{ij}, I) \wedge \text{North}(P_{ij}, Q) \Rightarrow \text{Bright}(Q, I),$$
$$\neg\text{Bright}(P_{ij}, I) \wedge \text{North}(P_{ij}, Q) \Rightarrow \neg\text{Bright}(Q, I),$$

where $\text{Bright}(P_{ij}, I)$ is the normalized brightness of pixel $P_{ij}$ in image $I$, and $\text{North}(P_{ij}, Q)$ indicates that $Q$ is the north neighbor of $P_{ij}$. We similarly include analogous rules for the south, east, and west neighbors, as well as the pixels mirrored across the horizontal and vertical axes. This setup results in up to 24 rules per pixel, which, in a 64 by 64 image, produces 80,896 weighted potential templates.

We train these HL-MRFs using MPE-approximate maximum likelihood with a 5.0 step size on the first 200 images of each data set and test on the last fifty. For training, we maximize the data log-likelihood of uniformly random held-out pixels for each training image, allowing for generalization throughout the image. Table 5 lists our results and others reported by Poon and Domingos [18]. HL-MRFs produce the best mean squared error on the left- and bottom-half settings for the Caltech101 set and the left-half setting in the Olivetti set. Only sum product networks produce lower error on the Olivetti bottom-half faces. Some reconstructed faces are displayed in Figure 1, where the shallow, pixel-based HL-MRFs produce comparably convincing images to sum-product networks, especially in the left-half setting, where HL-MRF can learn which pixels are likely to mimic their horizontal mirror. While neither method is particularly good at reconstructing the bottom half of faces, the qualitative difference between the deep SPN and the shallow HL-MRF reconstructions is that SPNs seem to hallucinate different faces, often with some artifacts, while HL-MRFs predict blurry shapes roughly the same pixel intensity as the observed, top half of the face. The tendency to better match pixel intensity helps HL-MRFs score better quantitatively on the Caltech101 faces, where the lighting conditions are more varied.

Table 5: Mean squared errors per pixel for image reconstruction. HL-MRFs produce the most accurate reconstructions on the Caltech101 and the left-half Olivetti faces, and only sum-product networks produce better reconstructions on Olivetti bottom-half faces. Scores for other methods are taken from Poon and Domingos [18].

|  | HL-MRF-Q (MLE) | SPN | DBM | DBN | PCA | NN |
| --- | --- | --- | --- | --- | --- | --- |
| Caltech-Left | 1741 | 1815 | 2998 | 4960 | 2851 | 2327 |
| Caltech-Bottom | 1910 | 1924 | 2656 | 3447 | 1944 | 2575 |
| Olivetti-Left | 927 | 942 | 1866 | 2386 | 1076 | 1527 |
| Olivetti-Bottom | 1226 | 918 | 2401 | 1931 | 1265 | 1793 |

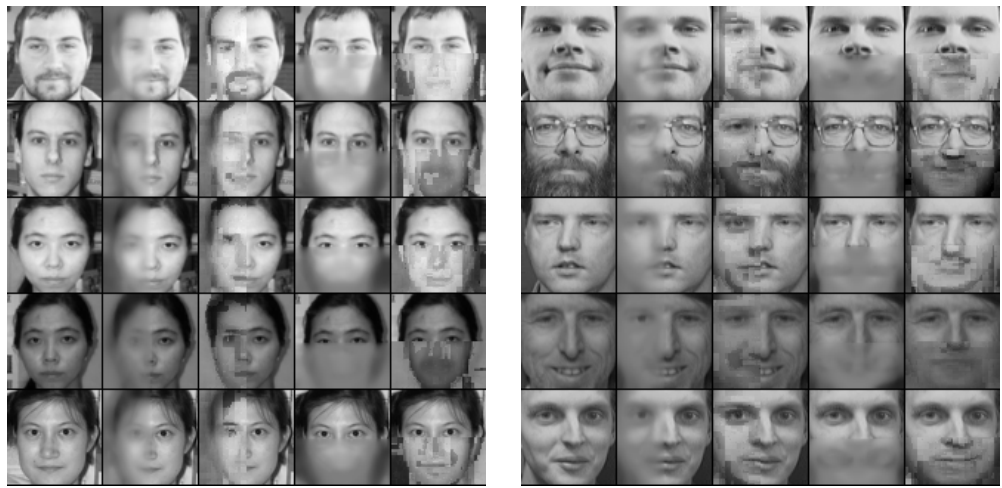

Figure 1: Example results on image reconstruction of Caltech101 (left) and Olivetti (right) faces. From left to right in each column: (1) true face, left side predictions by (2) HL-MRFs and (3) SPNs, and bottom half predictions by (4) HL-MRFs and (5) SPNs. SPN reconstructions are downloaded from Poon and Domingos [18].

Training and predicting with these HL-MRFs takes little time. In our experiments, training each model takes about 45 minutes on a 12-core machine, while predicting takes under a second per image. While Poon and Domingos [18] report faster training with SPNs, both HL-MRFs and SPNs clearly belong to a class of faster models when compared to DBNs and DBMs, which can take days to train on modern hardware.

## 6 CONCLUSION

We have shown that HL-MRFs are a flexible and interpretable class of models, capable of modeling a wide variety of domains. HL-MRFs admit fast, convex inference. The MPE inference algorithm we introduce is applicable to the full class of HL-MRFs. With this fast, general algorithm, we are the first to show results using quadratic HL-MRFs on real-world data. In our experiments, HL-MRFs match or exceed the predictive performance of state-of-the-art methods on four diverse tasks. The natural mapping between hinge-loss potentials and logic rules makes HL-MRFs easy to define and interpret.

**Acknowledgements**

This work was supported by NSF grants CCF0937094 and IIS1218488, and IARPA via DoI/NBC contract number D12PC00337. The U.S. Government is authorized to reproduce and distribute reprints for governmental purposes notwithstanding any copyright annotation thereon. Disclaimer: The views and conclusions contained herein are those of the authors and should not be interpreted as necessarily representing the official policies or endorsements, either expressed or implied, of IARPA, DoI/NBC, or the U.S. Government.